\pdfoutput=1

\documentclass[11pt]{article}

\usepackage{ACL2023}

\usepackage{times}
\usepackage{latexsym}
\usepackage{booktabs}
\usepackage{graphicx}

\usepackage[T1]{fontenc}

\usepackage[utf8]{inputenc}

\usepackage{microtype}

\usepackage{inconsolata}

\usepackage{hyperref}

\usepackage{listings}
\usepackage{comment}

\usepackage{caption}
\usepackage{subcaption}

\usepackage{siunitx}

\title{Understanding Programs by Exploiting (Fuzzing) Test Cases}

\author{
Jianyu Zhao\thanks{~~Equal contribution}\,\,$^{1}$, Yuyang Rong$^{\ast 2}$, Yiwen Guo\thanks{~~Corresponding author}\,\,$^3$, Yifeng He$^{2}$, Hao Chen$^2$  \\
$^1$Tencent Security Big Data Lab,\, $^2$UC Davis,\, $^3$Independent Researcher\\
\small{{yjjyzhao@tencent.com,\, \{PeterRong96,\,guoyiwen89\}@gmal.com}}\\
\small{{\{yfhe,\,chen\}@ucdavis.edu}}
}

\begin{document}
\maketitle
\begin{abstract}
    Semantic understanding of programs has attracted great attention in the community.
    Inspired by recent successes of large language models (LLMs) in natural language understanding, tremendous progress has been made by treating programming language as another sort of natural language and training LLMs on corpora of program code.
    However, programs are essentially different from texts after all, in a sense that they are normally heavily structured and syntax-strict.
    In particular, programs and their basic units (i.e., functions and subroutines) are designed to demonstrate a variety of behaviors and/or provide possible outputs, given different inputs.
    The relationship between inputs and possible outputs/behaviors represents the functions/subroutines and profiles the program as a whole.
    Therefore, we propose to incorporate such a relationship into learning, for achieving a deeper semantic understanding of programs.
    To obtain inputs that are representative enough to trigger the execution of most part of the code, we resort to fuzz testing and propose fuzz tuning to boost the performance of program understanding and code representation learning, given a pre-trained LLM.
    The effectiveness of the proposed method is verified on two program understanding tasks including code clone detection and code classification, and it outperforms current state-of-the-arts by large margins. Code is available at \url{https://github.com/rabbitjy/FuzzTuning}. 
\end{abstract}

\section{Introduction}
\label{sec:intro}

Code intelligence powered by machine learning has attracted considerable attention in both the AI and software engineering community. Particularly, code representation learning, which aims to encode functional semantics of source code, lays the foundation for achieving the intelligence and is of great interest. The learned representation can be applied to various downstream tasks, including code classification~\cite{mou2016convolutional}, code summarization~\cite{iyer2016summarizing}, clone detection~\cite{svajlenko2014towards, mou2016convolutional}, etc.

Many efforts inspired by the developments of natural language understanding have been devoted to learning code representations, among which it has been increasingly popular to adopt large language models (LLMs) that are capable of learning contextual information from data at scale~\cite{feng2020codebert, li2022competition}.
The LLMs can then be fine-tuned on domain-specific code to achieve superior performance compared with tradition models.

Despite being effective, these natural language processing methods do not fit perfectly for handling programs.
Specifically, programs are heavily structured and syntax-strict (to be understood by compilers or interpreters), while natural language corpus is not. 
As basic units of programs, functions and subroutines can take a variety of argument values to demonstrate different logical behaviors or return different results.
That being said, the relationship between inputs and possible outputs/behaviors essentially represents the functions/subroutines and further the whole programs.

In this paper, we propose to incorporate such a relationship into learning for a deeper understanding of programs.
In fact, given enough inputs to execute all pieces of the code, then the outputs would include enough runtime information we need to profile and understand the program.
However, it is nontrivial to generate a limited number of inputs that are representative enough to execute every part of the code.
Without a proper strategy, we may end up with a large number of inputs that execute similar parts of codes.
To address the issue, we opt to utilize fuzz testing (also known as fuzzing)~\cite{sutton2007fuzzing}, which is a common software testing practice and dynamic analysis tool whose original goal is to find software bugs by executing as much code as possible.
More specifically, we repurpose fuzz testing to generate input and output data for assisting code representation learning, and we demonstrate how the input and output data (i.e., test cases) can be appropriately incorporated into existing LLMs to achieve superior program understanding performance.

The contributions of this paper are three-fold.
First, by recognizing the essence code representation learning, we propose to take advantage of the relationship between inputs and possible outputs for achieving a deeper understanding of programs. Second, we, for the first time, repurpose fuzz testing to assist code presentation learning, marrying these two concepts from different communities for achieving more powerful AI. Third, we obtain state-of-the-art results on typical program understanding tasks including clone detection and code classification, in comparison to prior arts.

\section{Related Work}
\label{sec:related}

In this section, we introduce related work on code understanding (from the natural language understanding community) and fuzzing (from the software engineering community).

\subsection{Code Representation Learning}
\label{sec:code_represent}

Inspired by the success of LLMs in natural language processing~\cite{raffel2020exploring, liu2019roberta, devlin2018bert}, LLMs trained on programming languages have also been widely used to drive code intelligence.
For instance, \citet{kanade2020learning} proposed cuBERT to train BERT models on a curated and deduplicated corpus of 7.4M Python files from GitHub, and adapt the pre-trained models to various code classification tasks and a program repair task.
Thereafter, a bunch of methods have been developed and a variety of LLMs have been trained on code data, including CodeBERT~\cite{feng2020codebert}, CodeT5~\cite{wang2021codet5}, and CodeGPT~\cite{lu2021codexglue}.

The importance of comprehending syntax and structures for learning code representations has also been pointed out by several prior arts, and methods that incorporate programming-language-specified features, including abstract syntax tree~\cite{tipirneni2022structcoder, guo2022unixcoder}, control or data flow graphs~\cite{guo2020graphcodebert}, and intermediate representation of code~\cite{peng2021could} have been developed. These methods only utilize information available for static analysis.
It is generally difficult for static analysis to be both safe and sound~\citep{10.1145/3368826.3377927} when analyzing the behavior of programs.
For instance, a path that exists on the control flow graph may never be executed due to data-flow limitations.
Our work is the first to take dynamic program information (by generating and exploiting test cases with inputs and outputs) into account for code representation learning.
An input will lead to execution of part of the program and an output (or some behaviors if no output is required), which would reflect the functionality of that part of the program.
We hypothesize that if we have enough inputs to execute the code sufficiently, the outputs would also include enough runtime information that we need to profile the program.

\subsection{Fuzzing}
Fuzz testing, or fuzzing, is a process that tests the correctness of programs.
Fuzzing can be roughly considered as a four-stage loop.
First the program is executed with a given input.
Second, the behavior of the program is monitored to determine if any new behavior is triggered.
Third, if a new behavior is present, the corresponding input will be saved into a store, otherwise, the input is discarded as not interesting.
Finally, a mutator takes a saved input in the store, mutates it in different fashions and sends the input for another round of execution.

American Fuzzy Lop (AFL)\footnote{\url{https://lcamtuf.coredump.cx/afl/}} is the first fuzzer to implement behavior monitoring using branch coverage.
It tracks which edges of the control flow graph have been executed.
Since the invention of AFL, many innovations have been made to improve the overall fuzzing performance.
\citet{osti_10313742, collafl} modified branch coverage to lower the overhead while improving tracking sensitivity.
\citet{aflfast} proposed that power scheduling is better than a first-in-first-out queue for input store and improved the fuzzing performance by a magnitude.
\citet{angora} introduced new mutation algorithms and showed superior performance than random mutation.
Many of the changes have been incorporated into a more modern tool called AFL++~\citep{AFLplusplus}.

Unfortunately, current use of fuzzers only focuses on the bugs in the software~\cite{chen2018iotfuzzer, rong2020integrity, aschermann2019nautilus} and did not show possibility of adopting fuzzing results in code representation learning for AI. We identify that these results can be used to profile programs and improve the performance of code representation learning and program understanding.

\section{Method}
\label{sec:method}

As mentioned in \autoref{sec:intro}, programs show strict syntax.
To inspire deeper understanding of the syntax and logical behaviors of a program or functions/subroutines (which are the building blocks of the program), we attempt to exploit the relationship between their inputs and possible outputs/behaviors for achieving improved understanding of programs and code, akin to how engineers understands third-party code.

However, with existing learning techniques, it seems nontrivial to generate inputs that could lead to execution of sufficient part of the code, thus we resort to fuzzing to achieve this goal.

\subsection{Fuzzing for Obtaining Inputs (and Outputs)}
\label{sec:io}

Despite being widely adopted for testing software, fuzzing has rarely been adopted in machine learning tasks.
In general, fuzzing is a software testing practice, whose goal is to find software bugs by executing as much code as possible.
To achieve this, it executes the program with different inputs and monitors the behavior of each execution.
Therefore, as byproducts of fuzzing, a large number of inputs may be produced by a fuzzer, each triggering a new behavior of the program under test.

Fuzzing is programming language agnostic in general.
However, with only source code, we have to compile the programs into executable files for fuzzing.
We mainly describe details for four mainstream languages (C, C++, Java, and Python), and a tool was specifically designed to build the programs for fuzzing.
This tool interacts with the compiler or interpreter to automatically fix some problems that prevent it from being fuzzed.
Since the main aim of this work is to assist models to better understand programs, we fix problems that do not affect the semantics and functionality of code but prevent fuzzing.

For C and C++, we treat them as C++ files.
Some semantics-irrelevant errors in the program would prevent the code from compiling and fuzzing.
For example, missing headers, absent return of a \lstinline{main} function that is defined to have one, and misuse of reserved keyword.
In order to fix these compilation errors, we designed a compiler plugin that can automatically fix these.
First we run the lexer and parser on the program to gain abstract syntax tree, which would make code transformation much easier.
Then we designed a parser to parse error message from the compiler.
We introduce several fixes to correct the program for different errors.

\begin{enumerate}
    \item \textbf{Missing headers.} We included most commonly used headers in the C++ library at the head of each program.
    \item \textbf{Incorrect return type and/or arguments.} For instance, if a \lstinline{main} function is defined as ``\lstinline{int main()}'' but provides no return, we fixed it by added ``\lstinline{return 0;}'' to the end of the program. 
    \item \textbf{Misuse of keywords in the standard library.} Reserved keywords might be misused as variables and we added a ``\lstinline{fixed_}'' prefix to each of such variables to avoid keyword violation.
    \item \textbf{Incorrect \lstinline{struct} definition.} Many structures were defined without a semicolon after it, we will append that semicolon.
    \item \textbf{Undeclared identifier.} We notice that many programs use static values as a constant value, yet the value is sometimes missing. We would analyze the usage of the constants and insert definitions for them.
\end{enumerate}

For Java programs, we compiled them into bytecodes using Kelinci~\cite{kelinci} to instrument them for fuzzing.
Not all programs we tested were named \lstinline{Main.java} but they all defined a \lstinline{Main} class.
In order to compile them, we changed the \lstinline{Main} class to its file name in order to compile it.
For each program, a TCP server was added to communicate with a fork server which then sends data to the fuzzer.

Python is the most difficult language.
First many lexical errors are not as easy to fix as C/C++ and Java.
For example, if the program mixed tabs and spaces, it is hard to infer what is the intended indention.
To solve this, we used \texttt{autopep8}\footnote{\url{https://pypi.org/project/autopep8/}} to transform the program.
The next challenge is that Python2 and Python3 can't be easily distinguished, therefore, it is unclear which interpreter should be used for fuzzing.
To detect the version, also to verify the correctness of the transformation, we treated all code as Python3 in the first place and try to compile python program to bytecode using \texttt{py\_compile}.
If the compilation failed, then it was probably a Python2 implementation and we tried to convert it to Python3 using \texttt{2to3}\footnote{\url{https://docs.python.org/3/library/2to3.html}}.
Finally, we had to instrument the behavior monitoring and reporting to communicate with the fuzzer.
We used \texttt{py-afl-fuzz}\footnote{\url{https://pypi.org/project/python-afl/}} to achieve this.

We want to point out that all the changes made in this section are for fuzzing only.
When training models in the following sections, the programs remain unchanged.

We selected AFL++~\cite{AFLplusplus} as our fuzzer and fuzzed all experimental data on a server with 2 20-core 40-thread x86\_64 CPUs and 692GB of memory.
Each fuzzer only has one thread and ran until it exhausts all paths or a $K$-minute timeout is triggered.
The stored inputs that are of interest to AFL++ can then be utilized to execute the program and obtain outputs, and they constitute the fuzzing test cases.
The test cases (i.e., pairs of inputs and outputs) were produced in bytes and we may decode it into human readable strings.

\subsection{Model}
Although it is possible to train representation learning models from scratch using the obtained fuzzing test cases, it can be more effective to take advantage of previous pre-training effort.
In particular, given a pre-trained LLM, we attempt to take these test cases as model inputs somehow.
Considering that the LLM was mostly trained on programming language and natural language corpus~\cite{feng2020codebert, wang2021codet5, guo2022unixcoder}, the source code of the program is fed into the model together with fuzzing test cases, by concatenating the two parts.

\subsection{Prompting}
The fuzzing test cases in their raw format are a series of bytes, and, by decoding, we can obtain a series of Unicode strings which are unorganized.
To help LLMs better understand these test cases, we introduce cloze prompts~\cite{petroni2019language}.
Prompt have shown significant power in natural language processing since the invention of LLMs.
Considering that LLMs can be pre-trained on both natural language corpora and programming language corpora, we design both natural-language-based
prompts and programming-language-based prompts for each pair of input (denoted by \lstinline{[INPUT]}) and output (denoted by \lstinline{[OUTPUT]}) as follows:
\begin{enumerate}
    \item \textbf{Natural-language-based prompt:}
    
          (a) \lstinline{[SEP]} + ``input: '' + \lstinline{[INPUT]} + ``,'' + ``output: '' + \lstinline{[OUTPUT]};

          (b)  \lstinline{[SEP]} + ``input is '' + \lstinline{[INPUT]} + ``and'' + ``output is '' + \lstinline{[OUTPUT]};

    \item \textbf{Programming-language-based prompt:}

          (a)  \lstinline{[SEP]} + "\lstinline{cin>>}" + \lstinline{[INPUT]} + "\lstinline{;}" + "\lstinline{cout<<}" + \lstinline{[OUTPUT]}; (For C/C++)

          (b)  \lstinline{[SEP]} + ``\lstinline{System.in} '' + \lstinline{[INPUT]} + ``\lstinline{;}'' + ``\lstinline{System.out}'' + \lstinline{[OUTPUT]}; (For Java)

          (c)  \lstinline{[SEP]} + ``\lstinline{input()}'' + \lstinline{[INPUT]} + ``\verb|\|n'' + ``\lstinline{print}'' + \lstinline{[OUTPUT]}; (For Python)

\end{enumerate}

In experiments, we found that the programming-language-based prompts are more effective and we will stick with it in the sequel of this paper, if not specified.
This is unsurprising since the fuzzing test cases can stay in harmony with the source code with such a prompt.

Each prompted pair of input and output can be concatenated together before being further concatenated with the source code. Pre-trained LLMs can be tuned on downstream datasets with their inputs being modified to consider both the source code and fuzzing test cases.
We call this method \emph{fuzz tuning} in the paper.

\begin{table}[t]
    \centering
    \resizebox{0.8\columnwidth}{!}{
        \begin{tabular}{lcc}
            \toprule
            Dataset & {\# of problems} & {\# of programs} \\ 
            \midrule
            POJ104           & 104              & 52K              \\ 
            C++1000          & 1000             & 500K             \\ 
            Python800        & 800              & 240K             \\ 
            Java250          & 250              & 75K              \\ 
            \bottomrule
        \end{tabular}
    } \vskip -0.1in
    \caption{
        Dataset statistics.
    }
    \label{tbl:codenet-statistics} \vskip -0.1in
\end{table}

\section{Experimental Results}
\label{sec:exp}

In this section, we report experimental results to verify the effectiveness of our fuzz tuning.
We consider popular tasks (i.e., clone detection and code classification) and datasets involving mainstream languages including C, C++, Java, and Python. Experiments were performed on NVIDIA V100 GPUs using PyTorch 1.7.0~\cite{paszke2019pytorch} implementations.

\begin{table*}[t]
    \renewcommand{\arraystretch}{1.0}
    \centering
    \resizebox{1.4\columnwidth}{!}{
        \begin{tabular}{lccc}
            \toprule
            Method                                                 & Java250        & Python800      & C++1000$^\ast$ \\
            \midrule
            Rule-based w/SPT(AROMA)~\cite{puri2021project}        & 19.00          & 19.00          & -              \\
            GNN w/SPT(MISIM)~\cite{puri2021project}               & 64.00          & 65.00          & -              \\ 
            CodeBERT{\color{gray}+FineT}~\cite{feng2020codebert}  & 81.47          & 83.23          & 44.94          \\
            UniXcoder{\color{gray}+FineT}~\cite{guo2022unixcoder} & 84.35          & 85.00          & 49.75          \\
            \midrule
            CodeBERT{\color{blue}+FuzzT} (ours)                   & \textbf{83.39} & \textbf{85.64} & \textbf{54.92} \\
            UniXcoder{\color{blue}+FuzzT} (ours)                  & \textbf{86.74} & \textbf{86.01} & \textbf{60.21} \\
            \bottomrule
        \end{tabular}%
    }
    \caption{Clone detection results on CodeNet. Compared with normal fine-tuning (FineT), our fuzz tuning (FuzzT) leads to significant improvements and new state-of-the-arts. C++1000$^\ast$ contains 16\% of all problems, which is a roughly 6.3x downsample of the original dataset (see Table~\ref{tab:c1000clone} for results on other scales). Bold stats are better.}  
    \label{tab:clone-java} \vskip -0.1in
\end{table*}

\begin{table}[t]
    \renewcommand{\arraystretch}{1.0}
    \centering
    \resizebox{\columnwidth}{!}{
        \begin{tabular}{lc}
            \toprule
            Method                                                         & MAP@R          \\
            \midrule
            CodeBERT{\color{gray}+FineT}~\cite{feng2020codebert}          & 84.29          \\
            GraphCodeBERT{\color{gray}+FineT}~\cite{guo2020graphcodebert} & 85.16          \\
            PLBART{\color{gray}+FineT}~\cite{ahmad2021unified}            & 86.27          \\
            SYNCOBERT{\color{gray}+FineT}~\cite{wang2021syncobert}        & 88.24          \\
            CodeT5{\color{gray}+FineT}~\cite{wang2021codet5}              & 88.65          \\
            ContraBERT{\color{gray}+FineT}~\cite{liu2023contrabert}                        & 90.46          \\
            UniXcoder{\color{gray}+FineT}~\cite{guo2022unixcoder}         & 90.52          \\
            \midrule
            CodeBERT{\color{blue}+FuzzT} (ours)                           & \textbf{92.01} \\
            UniXcoder{\color{blue}+FuzzT} (ours)                          & \textbf{93.40} \\
            \bottomrule
        \end{tabular}
    } 
    \caption{Clone detection results on POJ104. Our fuzz tuning (FuzzT) leads to state-of-the-art results. Bold stats are better.} \vskip -0.2in
    \label{tab:clone-poj}
\end{table}

\textbf{Datasets.}
Our experiments were performed mainly on two datasets, i.e., POJ104~\citep{mou2016convolutional} and CodeNet~\citep{puri2021project}.
POJ104 has been incorporated into CodeXGlue~\cite{lu2021codexglue} and is widely used.
It consists of 104 problems, each containing 500 C/C++ implementations.
CodeNet is a recently proposed large-scale dataset for AI for code applications, and it contains programs written in C++, Java, and Python.
In particular, it has four subsets for these languages: Java250, Python800, C++1000, and C++1400.
We chose Java250, Python800, and C++1000 for experiments, which cover all the three languages in CodeNet.
Java250 consists of 250 problems where each includes 300 Java programs, and Python800 consists of 800 problems where each includes 300 Python programs. C++1000 consists of 1000 problems where each includes 500 C++ programs, and it is mainly used to verify the effectiveness of our method over various training scales (i.e., our fuzz tuning will be performed on various subsampling ratios of the set) in this paper.
See \autoref{tbl:codenet-statistics} for a summarization of key information of all datasets.

\textbf{Pre-trained LLMs. }
To make our experiments more comprehensive, our fuzz tuning (FuzzT) was tested on two different LLMs: CodeBERT~\cite{feng2020codebert} and UniXcoder~\cite{guo2022unixcoder} that were pre-trained on both natural languages and programming languages.

\textbf{Obtaining test cases. }
We set $K=5$ for the fuzzer.
In POJ104, 90.3\% of all fuzzed programs quits before timeout, which justifies our decision.
Taking advantage of our effort in \autoref{sec:io}, all datasets have more than 95\% of the programs compiled / validated, and all of them have more than 90\% of programs fuzzed. 

\begin{table*}[t]
    \renewcommand{\arraystretch}{1.0}
    \centering
    \resizebox{1.4\columnwidth}{!}{%
        \begin{tabular}{lccc}
            \toprule
            Method                                                 & Java250         & Python800       & C++1000$^\dagger$ \\
            \midrule
            GIN~\cite{puri2021project}                            & 6.74\%          & 5.83\%          & -                 \\
            CodeGraph+GCN~\cite{puri2021project}                  & 5.90\%          & 12.2\%          & -                 \\
            C-BERT{\color{gray}+FineT}~\cite{puri2021project}     & 2.60\%          & 2.91\%          & -                 \\
            
            CodeBERT{\color{gray}+FineT}~\cite{feng2020codebert}  & 2.37\%          & 2.19\%          & 16.34\%           \\ 
            UniXcoder{\color{gray}+FineT}~\cite{guo2022unixcoder} & 2.02\%          & 1.95\%          & 14.57\%           \\ 
            \midrule
            CodeBERT{\color{blue}+FuzzT} (ours)                   & \textbf{1.77\%} & \textbf{1.61\%} & \textbf{7.63\%}   \\
            UniXcoder{\color{blue}+FuzzT} (ours)                  & \textbf{1.56\%} & \textbf{1.28\%} & \textbf{6.18\%}   \\
            \bottomrule
        \end{tabular}
    }
    \caption{Code classification results on CodeNet. Our fuzz tuning (FuzzT) leads to new state-of-the-arts. C++1000$^\dagger$ contains 40\% of all problems, which is a 2.5x downsample of the original dataset (see Table~\ref{tab:c1000cls} for results on other data scales). Bold stats are better.} 
    \label{tab:classify} \vskip -0.1in
\end{table*}

\begin{table}[t]
    \renewcommand{\arraystretch}{1.0}
    \centering
    \resizebox{\columnwidth}{!}{%
        \begin{tabular}{lc}
            \toprule
            Method                                                 & Error Rate      \\
            \midrule
            TBCNN~\cite{mou2016convolutional}                     & 6.00\%          \\
            ProGraML~\cite{cummins2020programl}                   & 3.38\%          \\
            OSCAR~\cite{peng2021could}                            & 1.92\%          \\
            CodeBERT{\color{gray}+FineT}~\cite{feng2020codebert}  & 1.61\%          \\
            UniXcoder{\color{gray}+FineT}~\cite{guo2022unixcoder} & 1.61\%          \\
            \midrule
            CodeBERT{\color{blue}+FuzzT} (ours)                   & \textbf{1.40\%} \\
            UniXcoder{\color{blue}+FuzzT} (ours)                  & \textbf{1.38\%} \\
            \bottomrule
        \end{tabular}%
    }
    \caption{Code classification results on POJ104. Our fuzz tuning (FuzzT) leads to new state-of-the-arts. Bold stats are better.} \vskip -0.20in
    \label{tab:classify-poj}
\end{table}

\subsection{Clone Detection Results}
\label{sec:ccd_results}
The clone detection task aims to measure the similarity between two code snippets or two programs, which can help reduce bugs and prevent the loss of intellectual property.

Given a code snippet or the source code of a program as a query, models should detect semantically similar implementations on the test set.
POJ104 is adopted as the default dataset for code clone detection on CodeXGlue, thus we did experiment on it first.
We followed previous work~\cite{lu2021codexglue} and used a 64/16/24 split.
That said, training was performed on 64 problems while validation and test were performed on other 16 and 24 problems, respectively.
Besides, We further experimented on CodeNet which shows a larger data scale and variety.
Java250, Python800, and C++1000 were used, and we followed~\citet{puri2021project} which used a 50\%/25\%/25\% split for training/validation/test for all these concerned sets.
C++1000 is mainly used to test our method over various training scales in \autoref{sec:scale} with the full test set, and we will only discuss results on the smallest subsampling ratio (which is roughly 6.3x, achieving by randomly selecting 16\% of the problems for experiments) in this subsection. We denoted such a subsampled set as C++1000$^\ast$. All models tested here were tuned on a single V100, for no longer than 8 GPU hours.

Results were evaluated using the mean average precision@R (MAP@R)~\citep{musgrave2020metric}.
To train our models, we followed previous work~\cite{lu2021codexglue} and directly set the learning rate, batch size, and maximal sequence length (for code tokens) to 2e-5, 8, and 400, respectively. We used the Adam optimizer~\cite{kingma2014adam} to fine-tune each pre-trained model for 2 epochs. The best model on the validation set is selected to test. We adopted the same hyper-parameters on both POJ104 and CodeNet.

\autoref{tab:clone-poj} provides the results on POJ104.
It is obvious that, when the proposed fuzz tuning is applied, we obtain significant performance gains with both CodeBERT and UniXcoder.
More specifically, comparing with the normal fine-turning (FineT) method, we obtained \textbf{+7.72\%} and \textbf{+2.88\%} absolute gains in MAP@R with CodeBERT and UniXcoder, respectively.
Such practical improvements clearly demonstrate the benefits of incorporating fuzzing test cases into program understanding.
In addition, fuzz tuning obtained models (i.e., CodeBERT+FuzzT and UniXcoder+FuzzT) outperform all other state-of-the-art models significantly, leading to obvious empirical superiority (i.e., \textbf{+2.94\%}) comparing even with a very recent winning solution on the CodeXGLUE benchmark \footnote{\url{https://microsoft.github.io/CodeXGLUE/}} called ContraBERT.

\autoref{tab:clone-java} demonstrates the results on CodeNet.
Apparently, on CodeNet, our fuzz tuning also leads to significant performance gains on CodeBERT and UniXcoder, when compared with FineT.

\subsection{Code Classification Results}
\label{sec:ac_results}
The concerned code classification task~\cite{mou2016convolutional} requires that we assign the same label to programs that were implemented to solve the same problem and achieve the same goal.
The experiments were also performed on POJ104 and CodeNet, where each unique problem is considered as a single class.
For POJ104, we adopted the same dataset split as in~\citet{peng2021could}'s work, and, for CodeNet, we kept the official data split~\cite{puri2021project}.
As previously mentioned, C++1000 in CodeNet is mainly used to test our method over various training scales in \autoref{sec:scale}, and we will only discuss results on the smallest subsampling ratio (which is 2.5x, achieving by randomly selecting 40\% of the programs for experiments) here. We denoted such a subsampled tuning set as C++1000$^\dagger$.
We followed the hyper-parameter setting of CodeBERT in defect detection to set a learning rate of 2e-5, a training batch size of 32, and a maximal sequence length (for code tokens) of 400.
We tuned pre-trained LLMs for 10 epochs and selected the models that performed the best on the validation set and report their results on the test sets. Error rate of different methods are reported for comparison. All our code classification models were were tuned on two V100, for no longer than 8 hours. 

\autoref{tab:classify-poj} and \autoref{tab:classify} summarize the results. Similarly, We observe that our fuzz tuning bring significant improvement, comparing with the normal fine-tuning (FineT) method, it leads to \textbf{+0.21\%} and \textbf{+0.23\%} absolute performance gain in reducing the error rate with CodeBERT and UniXcoder, respectively, on POJ104. Fuzz tuning obtained models (i.e, CodeBERT+FuzzT and UniXcoder+FuzzT) also outperform all previous models on this task on POJ104, leading to new state-of-the-arts.
The same conclusion can also be drawn on CodeNet, showing that the effectiveness of our method hold on various programming languages.

The results on both clone detection and code classification demonstrate the effectiveness of our fuzz tuning.
Both tasks requires the model to understand not only the structure of the code, but further the semantics, which is hard to acquire by simply looking at the code.
Yet, provided with inputs and outputs, the model can excel.
We contribute this accuracy gain to program profiling provided through fuzzing.
These profiles include essential dynamic information that isn't used by any other models.

\subsection{Ablation Study}
\label{sec:ablation}

In this subsection, we investigate impacting factors in our method: including the quality of test cases, decoding, and prompting.

\begin{table}[t]
    \renewcommand{\arraystretch}{1.0}
    \centering
    \resizebox{0.7\columnwidth}{!}{
        \begin{tabular}{lcc}
            \toprule
            Decoding                       & CD             & CC              \\
            \midrule
            CodeBERT{\color{gray}+FineT} & 84.29          & 1.61\%          \\
            CodeBERT{\color{blue}+FuzzT}                              \\
            ~~\small{-in bytes}          & 84.21          & 1.55\%          \\
            ~~\small{-in UTF-8 string}   & \textbf{92.01} & \textbf{1.40\%} \\
            \bottomrule
        \end{tabular} }
    \caption{Comparing using raw and decoded fuzzing test cases in tuning clone detection (CD) and code classification (CC) models on POJ104. MAP@R and the error rate are evaluated for the two tasks, respectively. Bold stats are better.} \vskip -0.05in
    \label{tab:ablation}
\end{table}

\begin{table}[t]
    \renewcommand{\arraystretch}{1.0}
    \centering
    \resizebox{0.7\columnwidth}{!}{
        \begin{tabular}{lcc}
            \toprule
            Prompt Type                     & CD             & CC              \\
            \midrule
            CodeBERT{\color{gray}+FineT}    & 84.29          & 1.61\%          \\
            CodeBERT{\color{blue}+FuzzT}                                \\
            ~~\small{-w/o prompt}           & 89.36          & 1.51\%          \\
            ~~\small{-NL prompt, type (a)}  & 91.14          & 1.54\%          \\
            ~~\small{-NL prompt, type (b)}  & 91.59          & 1.58\%          \\
            ~~\small{-PL prompt, for C/C++} & \textbf{92.01} & \textbf{1.40\%} \\
            \bottomrule
        \end{tabular} }
    \caption{Comparing different prompts for our fuzz tuning on the clone detection (CD) and code classification (CC) tasks on POJ104. Bold stats are better.} \vskip -0.15in
    \label{tab:prompt}
\end{table} 

\textbf{Random cases vs fuzzing cases. }
Given the success of fuzz tuning in clone detection and code classification, the effectiveness of incorporating test cases can be recognized.
One may expect that random input generator can work to some extent, for providing test cases.
Unfortunately, our evaluation shows otherwise.
We tried following this idea and crafted around 2000 inputs for each program, yet none of them is valid and understandable to the program.
This result is expected, since the chance of a byte being a digit is only 10/256, there is less than 1\% change of generating a 3-digit number.
Thus, it is reasonable to conclude that random input generator is prone to generating invalid inputs, which lead to crash and hang of the program and cannot be used to profile it.
By contrast, our fuzzer provides behavior monitoring, all these ineffective inputs are filtered and not reported in the first place.

\textbf{Decoding. }
As mentioned in \autoref{sec:io}, the fuzzer processes obtained inputs as a series of bytes. We argue that reading test cases as bytes will cause severe performance degradation, since LLMs are pre-trained using human-readable codes and natural languages, which explains why we decode the obtained bytes before feeding them to LLMs.
To verify the effectiveness of decoding, we compare using human-readable UTF-8 strings and those raw bytes, both with cloze prompts, for program understanding.
The experiment was conducted on the POJ104 clone detection task and the POJ104 code classification task. 
\autoref{tab:ablation} shows the results. Apparently, human-readable test cases perform much better than bytes-format ones on both two tasks.

\textbf{Prompting. }
We then compare the performance of fuzz tuning with and without prompting. \autoref{tab:prompt} demonstrates the results. For prompting, two types of natural-language-based (NL-based) prompts and the advocated programming-language-based (PL-based) prompt are tested.
Apparently, prompting is beneficial.
As has been mentioned in \autoref{sec:method}, the PL-based prompt outperforms the two types of NL-based prompts. It shows a \textbf{+2.65\%} absolute gain on the POJ104 clone detection task and a \textbf{+0.11\%} absolute gain on code classification, compared with an implementation of fuzz tuning without prompts. For clone detection, prompting is always effective, no matter it is NL-based or PL-based, while, for code classification, the NL-based prompts fail.

\begin{table}[t]
    \resizebox{\columnwidth}{!}{
        \begin{tabular}{lccc}
            \toprule
            Method                        & 4\% ($\downarrow$25x) \  & 8\% ($\downarrow$12.5x) & 16\% ($\downarrow\sim$6.3x) \\
            \midrule
            
            CodeBERT{\color{gray}+FineT}  & 26.92                  & 36.79                 & 44.94                     \\
            CodeBERT{\color{blue}+FuzzT}  & \textbf{30.66}         & \textbf{40.57}        & \textbf{54.92}            \\
            UniXcoder{\color{gray}+FineT} & 34.71                 & 43.06                 & 49.75                    \\
            UniXcoder{\color{blue}+FuzzT} & \textbf{42.53}         & \textbf{51.83}        & \textbf{60.21}            \\
            \bottomrule
        \end{tabular}
    }
    \caption{How different methods scale with the size of training/tuning dataset on the C++1000 \emph{clone detection} task. Bold stats are better.} \vskip -0.1in
    \label{tab:c1000clone}
\end{table}

\begin{table}[t]
    \resizebox{\columnwidth}{!}{%
        \begin{tabular}{lccc}
            \toprule
            Method                        & 10\% ($\downarrow$10x) & 20\% ($\downarrow$5x) \ \  & 40\% ($\downarrow$2.5x)\ \\
            \midrule
            CodeBERT{\color{gray}+FineT}  & 34.38\%                & 19.15\%                    & 16.34\%                  \\
            CodeBERT{\color{blue}+FuzzT}  & \textbf{30.90\%}       & \textbf{12.53\%}           & \textbf{7.63\%}         \\
            UniXcoder{\color{gray}+FineT} & 21.66\%                & 16.24\%                    & 14.57\%                  \\
            UniXcoder{\color{blue}+FuzzT} & \textbf{14.48\%}       & \textbf{8.39\%}           & \textbf{6.18\%}         \\
            \bottomrule
        \end{tabular}%
    }
    \caption{How different methods scale with the size of training/tuning dataset on the C++1000 \emph{code classification} task. Bold stats are better.} \vskip -0.1in
    \label{tab:c1000cls}
\end{table}

\subsection{Data Scale}
\label{sec:scale}
We then investigate whether our fuzz tuning is effective on various training data scales.
To achieve this, we subsampled from C++1000 in CodeNet to construct data sets of various scales to perform fuzz tuning.
The official split of C++1000 was considered to construct test sets~\cite{puri2021project}, and the same test sets were adopted for testing models obtained on all these training scales.
For clone detection, we sampled 4\%, 8\%, and 16\% of the training and validation problems (i.e., subsampled the training and validation set by 25x, 12.5x, and roughly 6.3x).
For the code classification task, we sampled 10\%, 20\% and 40\% (i.e., subsampled by 10x, 5x, and 2.5x) of the training and validation and keep the sample ratio between the two sets as 4:1.
We adopted the same experimental settings as in \autoref{sec:ccd_results} and \autoref{sec:ac_results}.

Results of different fuzz tuning scales are provided in~\autoref{tab:c1000clone} and~\autoref{tab:c1000cls}. Apparently, our fuzz tuning is effective on all these training scales.
In particular, for clone detection, when only 4\% of the data is used for training, normal fine-tuning of CodeBERT and UniXcoder shows an MAP@R of only 26.92 and 34.71, respectively, while, by introducing fuzz tuning, we can achieve 30.66 and 42.53, respectively, showing even more obvious superiority than with 16\% of the data. It is also possible for both fine-tuning and fuzz tuning to scale to more than 16\% of the data, yet it requires more than 10 epochs to reach their performance plateaus and weaken the necessity of pre-training, thus we will leave it to future work for exploration. The same conclusion can be drawn for code classification.

\subsection{Case Study}
\begin{figure}[t]
    \centering
    \includegraphics[width=\linewidth]{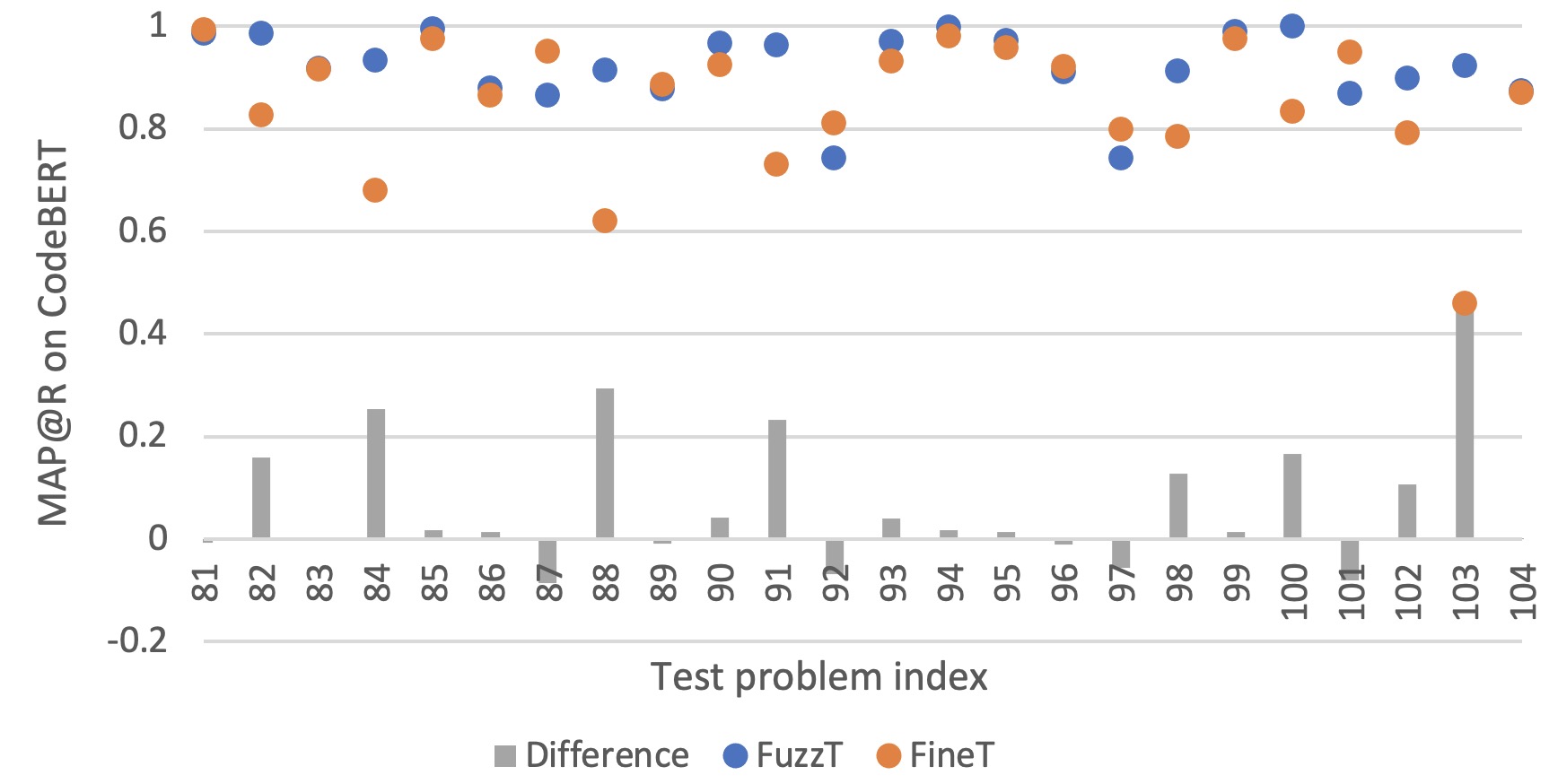} \vskip -0.12in
    \caption{Per-problem clone detection performance on the POJ104 test set, using \emph{CoderBERT+FineT} or \emph{CoderBERT+FuzzT}. The horizontal axis shows the ID of the POJ104 problems, and the vertical axis is the MAP@R.} \vskip -0.22in
    \label{fig:case-study}
\end{figure}

In this section, we extract some real cases in the concerned dataset (i.e., POJ104) to show how our fuzz tuning works.
\autoref{fig:case-study} reports the achieved per-program MAP@R and the performance gap between FuzzT and FineT on the POJ104 test set, \emph{with CodeBERT}.
We see that FuzzT outperforms FineT on 17 out of the 24 test problems.

\begin{figure}[h]
    \centering\includegraphics[width=\linewidth]{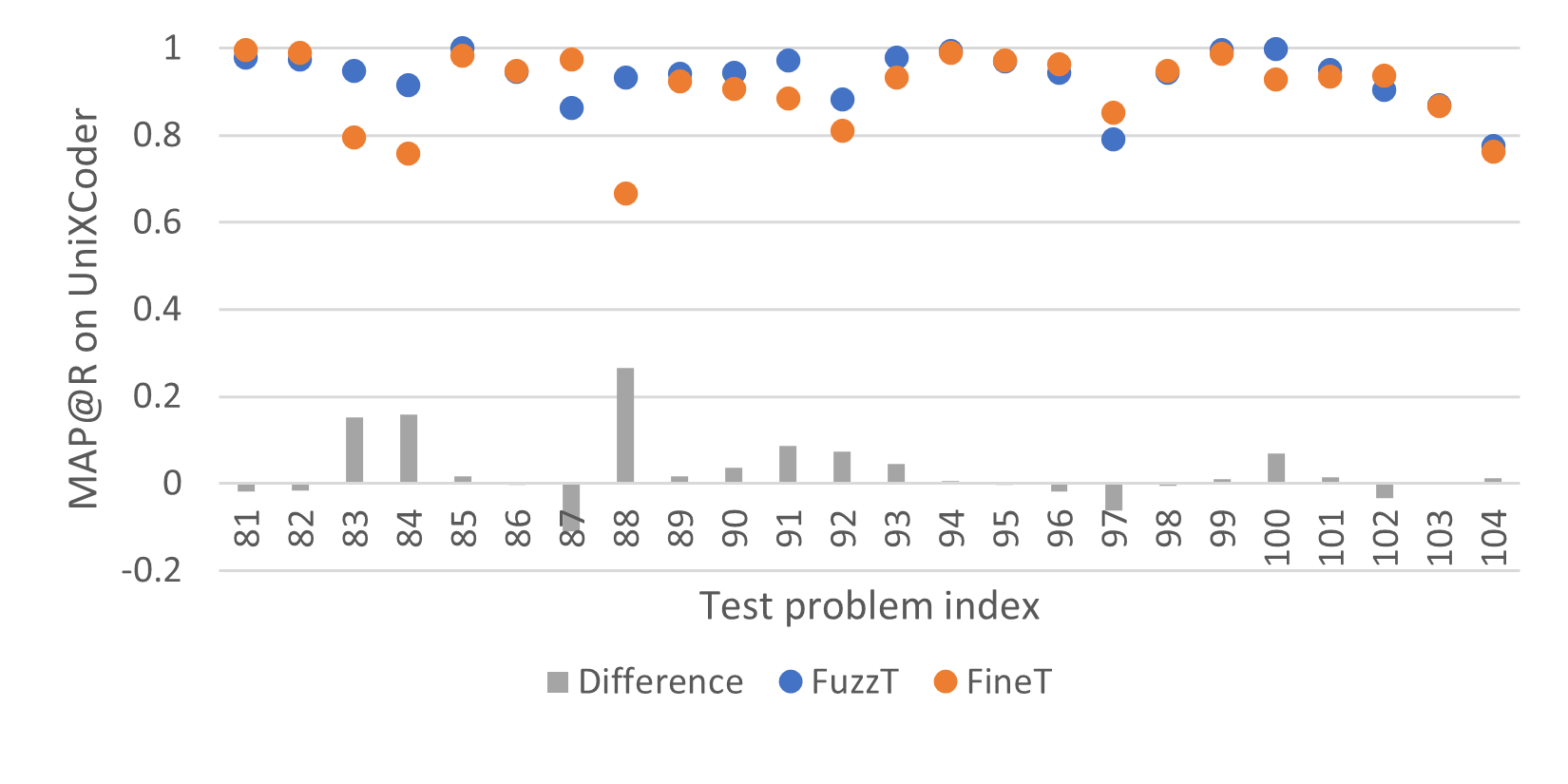} \vskip -0.15in
    \caption{Per-problem clone detection performance on the POJ104 test set, using \emph{UniXcoder+FineT} or \emph{UniXcoder+FuzzT}. The horizontal axis shows the ID of the POJ104 problems, and the vertical axis is the MAP@R.} \vskip -0.15in
    \label{fig:appendix}
\end{figure}

\autoref{fig:case-study} demonstrates that using the normal fine-tuning leads to very low MAP@R on Problem 103 of POJ104~\footnote{Note that Problem 1-80 are training and validation problems, and Problem 81-104 are test problems.}, yet our fuzz tuning more than doubled the score.
Although POJ104 does not describe each problem in detail, we did some investigations and conjecture that this particular problem is asking how many identical consecutive letters are there in a given string, if letter case is ignored.
Our investigations show that many programmers all unifies the letter case in the string first, but they disagree on whether to use uppercase letters or lowercase letter sand disagree on how to achieve this, leading to different implementations including utilizing standard library calls (i.e., to convert each character ``c'' using ``\lstinline{toupper(c)}''), calculating offset by casting (i.e., implementing something like \lstinline{c-'A'+'a'}), and static mapping (i.e., using ``\lstinline{caseMap[c]}'').
This will pose challenges to models for understanding their functionality, if fine-tuning on source code only.
One may expect this particular problem to be addressed by pre-training on relevant data or by taking more advantage of static information of programs. This is possible, since for other pre-trained LLMs, Problem 103 may not be the most challenging one. However, other issues similarly exist, e.g., UniXcoder+FineT shows its worst performance on Problem 88, as can be seen in \autoref{fig:appendix}.

\vskip -0.1in

By contrast, since the programs achieve the same goal, the test cases can help convey that information to the model. This further demonstrates our idea and explains the effectiveness of our fuzz tuning.

\section{Conclusion}
In this paper, we have pointed out that exploiting informative test cases helps to understanding of programs.
We have developed fuzz tuning, a novel method which takes advantage of fuzzing together with prior large-scale pre-training effort to achieve this goal.
Our fuzz tuning repurposes traditional fuzzers to generate informative test cases that well-represent the functionality of programs and it introduces appropriate cloze prompts to incorporate the test cases into being processed.
By performing comprehensive experiments on two datasets and two program understanding tasks, we have verified the effectiveness of the method and achieved new state-of-the-arts.

\section*{Limitations}
Fuzzers are designed to reach deep and complex control flow in large software.
Many programs for current AI for code datasets do not have complex control flow.
As a result, AFL++ can quickly cover all program branches before generating many inputs for us to feed to the model.
We plan to try data-flow coverage as a more accurate coverage metric in the future.

AFL++ uses branch coverage to track fuzzing progress.
Although it works well on C/C++ programs, it may be ineffective on languages with exceptions, which are implicit control flow.
For example, AFL++ cannot distinguish different exceptions thrown in the same block, which sometimes leads to low coverage in Python programs.
To overcome this issue, one possible way is to change from branch coverage to line coverage.

Although our current implementation requires a fuzzer, our approach can also work on tasks with only functions or code snippets as long as we can acquire adequate input/output pairs of the functions or code snippets, which may have some engineering challenges but is not infeasible. For example, in recent years, the software engineering community has proposed various ways to fuzz bare functions~\cite{serebryany2016continuous, ispoglou2020fuzzgen}.

\section*{Ethical Consideration}
Our method exploits fuzzing test cases for program understanding. 
Improved semantic understanding of programs facilitates various tasks, e.g., code generation and code completion, which might further be used to patch vulnerabilities or fix defects of softwares and systems. 
Nevertheless, considerable effort has to be further devoted to apply the method to these applications, for which we encourage to take special care in advance.
In addition, a number of crashes and hangs have been observed on programs in the adopted datasets, since fuzz testing is utilized.
We do not demonstrate test cases that lead to these crashes and hangs to avoid misuse of this information. 

\section*{Acknowledgment}

This material is partially based upon work supported by the National Science Foundation under Grant No.\ 1801751 and 1956364.

\bibliography{custom}
\bibliographystyle{acl_natbib}

\end{document}